\documentclass[sn-mathphys,Numbered]{sn-jnl}

\usepackage{graphicx}%
\usepackage{multirow}%
\usepackage{amsmath,amssymb,amsfonts}%
\usepackage{amsthm}%
\usepackage{mathrsfs}%
\usepackage[title]{appendix}%
\usepackage{xcolor}%
\usepackage{textcomp}%
\usepackage{manyfoot}%
\usepackage{booktabs}%
\usepackage{algorithm}%
\usepackage{algorithmicx}%
\usepackage{algpseudocode}%
\usepackage{listings}%

\theoremstyle{thmstyleone}%
\theoremstyle{thmstyletwo}%

\theoremstyle{thmstylethree}%

\begin{document}

\title[Article Title]{FATURA: A Multi-Layout Invoice Image Dataset for Document Analysis and Understanding}


\author[1]{\fnm{Mahmoud} \sur{  Limam}}\email{mahmoud.limem@enetcom.u-sfax.tn}
\author*[1,2]{\fnm{Marwa} \sur{  Dhiaf}}\email{marwa.dhiaf.doc@enetcom.usf.tn}

\author[1,2]{\fnm{Yousri} \sur{  Kessentini}}\email{yousri.kessentini@crns.rnrt.tn}

\affil*[1]{\orgdiv{Digital Research Center of Sfax, 3021, Sfax, Tunisia}, \orgname{Organization}, \country{Tunisia}}

\affil[2]{\orgdiv{SM@RTS : Laboratory of Signals, systeMs, aRtificial Intelligence and neTworkS, Sfax, Tunisia}}




\abstract{Document analysis and understanding models often require extensive annotated data to be trained. However, various document-related tasks extend beyond mere text transcription, requiring both textual content and precise bounding-box annotations to identify different document elements. Collecting such data becomes particularly challenging, especially in the context of invoices, where privacy concerns add an additional layer of complexity.
In this paper, we introduce FATURA, a pivotal resource for researchers in the field of document analysis and understanding. FATURA is a highly diverse dataset featuring multi-layout, annotated invoice document images. Comprising $10,000$ invoices with $50$ distinct layouts, it represents the largest openly accessible image dataset of invoice documents known to date. We also provide comprehensive benchmarks for various document analysis and understanding tasks and conduct experiments under diverse training and evaluation scenarios. The dataset is freely accessible at this \href{https://zenodo.org/record/8261508}{URL\footnote{https://zenodo.org/record/8261508}}, empowering researchers to advance the field of document analysis and understanding. }
\keywords{Open invoice images dataset  ,Document layout analysis, Key-value extraction, Document Layout analysis}

\maketitle
\section{Introduction}\label{sec1}

In an age dominated by digital information, document analysis and understanding are fundamental pillars of information retrieval and knowledge extraction. The ability to automatically process, interpret, and extract insights from documents has far-reaching applications across various domains, including finance, healthcare, legal, and administrative sectors.  Invoices are one of the most prevalent and important types of documents in this context, as they represent a significant portion of the financial transactions that occur between businesses.

Invoices contain a wealth of information that can provide valuable insight into business operations, including the names of the parties involved, product descriptions, and financial information. This information is essential for tracking sales, managing inventory, and monitoring cash flow. However, analyzing and understanding invoices can be challenging due to the variability and complexity of these documents. For example, invoices from different industries come in a variety of formats and layouts, often containing sensitive data. The task of collecting and annotating a diverse set of invoice documents is not only labor intensive, but is also complicated by privacy concerns. Balancing the need for data diversity with the imperative of safeguarding privacy adds an additional layer of complexity to this endeavor.

Document analysis and understanding models are heavily relying on annotated datasets for effective training. While basic text transcription suffices for some tasks, many document-related challenges require a deeper level of annotation. This entails not only recognizing textual content but also precisely delineating the boundaries of different elements within a document through bounding-box annotations. These annotations facilitate the extraction of structured information, making it possible to discern crucial component of the document, such as tables, titles, and signatures.

To address these challenges, researchers have turned to generative models, such as generative adversarial networks (GANs)~\cite{ref_gan,ref_vqgan,ref_dalle2}, as potential solutions for generating invoices. However, the effectiveness of generative models in generating unstructured documents, such as invoices, is limited by the lack of a clear and consistent structure in these documents.

On the contrary, deep learning architectures have demonstrated remarkable efficacy in various document-related tasks. Some of these models were originally designed to process image data broadly, but have exhibited exceptional performance when applied to structured document images. This category includes Convolutional Neural Networks (CNNs) \cite{ref_model13, ref_model1}, and Vision Transformer (ViT)~\cite{ref_model12}.

Conversely, there are models purposefully crafted for handling document images, often adopting a multi-modal approach. These models not only analyze the visual aspects of the image, but also dive into the textual content and even the layout information. An illustrative instance is the LayoutLM architectures \cite{ref_model6,ref_model7,ref_model8}.

Nevertheless, it is crucial to note that these systems frequently rely on cutting-edge deep learning models. Deep learning models, renowned for their data-hungry nature, require substantial amounts of high-quality data for effective training. Consequently, the growing demand for robust systems and models underscores the need for large, well-annotated datasets. Acquiring such datasets for invoice processing proves to be a formidable and time-consuming endeavor, as it mandates extensive manual effort to extract and annotate pertinent information from invoices.

Current datasets for invoice recognition, like ~\cite{ref_data3,ref_data6}, are encumbered with various limitations. These restrictions pertain to both the quantity of samples and the diversity of invoice layouts, thereby impeding their capacity to effectively assess the performance of text recognition algorithms. Moreover, certain publicly available datasets may not contain the level of annotation required for specific tasks. Some datasets, for instance, may solely support document classification, lacking the essential bounding-box coordinates and textual details required for more comprehensive analysis. Furthermore, even when a high-quality dataset of document images is accessible, it might fall short in terms of quantity, particularly when targeting a specific document type like invoices. This shortfall presents challenges, as certain tasks demand a robust supply of annotated document images of a particular category.

In light of these challenges, we introduce FATURA, an exemplary contribution to the realm of document analysis and understanding. FATURA constitutes a groundbreaking dataset meticulously crafted to address the prevailing limitations. This monumental dataset comprises a staggering total of $10,000$ invoice document images, each adorned with one of $50$ unique layouts, making it the most extensive openly accessible invoice document image dataset. This resource caters to the needs of researchers by not only offering diversity in data but also presenting an extensive benchmark for various document analysis tasks. Our aim is to provide the research community with an indispensable tool, empowering them to advance the field of document analysis and understanding while respecting the ever-important considerations of data privacy.

In the following sections, we start by providing an in-depth review of related works, where we introduce various document understanding approaches and examine the landscape of existing datasets in the field. Next, we explore the core aspects of FATURA, providing detailed insights of its content, how it is annotated, and the specific challenges it aims to overcome. Additionally, we conduct an evaluation of cutting-edge methods designed for the automatic extraction of named entities from invoices under diverse training and evaluation scenarios.

\section{Related work}

In the rapidly evolving field of document analysis and understanding, researchers have explored various systems, methodologies, and datasets to improve the automation of document processing tasks. This section delves into the extensive body of related work, starting with an investigation into different systems and methodologies for document understanding. We then dedicate the second subsection to a comprehensive exploration of existing invoice datasets.

\subsection{Document Understanding Approaches}
Document understanding encompasses a wide array of tasks, including layout analysis, information extraction, named entity recognition, and document image classification. Numerous systems and methodologies have been developed to address these challenges. This research field has witnessed significant advancements, driven primarily by deep learning techniques. These approaches have revolutionized the field by achieving remarkable results in a wide range of document analysis tasks. Various architectural paradigms, spanning image-based, text-based, and multi-modal models, have contributed to these successes.

Image-based architectures, exploit the success of deep CNN and object detection models \cite{ijdar} to decompose the document image into semantically meaningful regions such as tables, graphical elements, titles, paragraphs...   Such models excel at capturing intricate patterns, textures, and spatial relationships within visual content. For instance, in \cite{ICSTCC}, authors consider the task of information extraction from invoices as an object detection task. To this end, they used three different models YOLOv5, Scaled YOLOv4 and Faster R-CNN ~\cite{ref_model2} to detect key field information in invoices. Additionally, they propose a data preprocessing method that helps to better generalize the learning. The authors in \cite{ijdar} focused on going beyond object detection to understand document layouts. They propose an instance segmentation model which, in contrast to classical document object detection pipelines, provides a more fine-grained instance-level segmentation for every individual object category of a document image. 
Image-based approaches relying on object detection and semantic segmentation have a notable limitation; they predominantly focus on visual document aspects while potentially neglecting critical textual information. They may not inherently understand the arrangement of text, the significance of specific text regions, or the semantic context of the document's content. This limitation becomes especially prominent in complex documents with varied layouts and fonts.


Text-based approaches involve the conversion of a document image into a textual representation, with the choice between Optical Character Recognition (OCR) or Handwritten Text Recognition (HTR) depending on the nature of the document (printed or handwritten). Subsequently, sophisticated natural language processing techniques are applied to parse the resulting text and extract the named entity tags \cite{iconip}. One significant weakness of this approach lies in its vulnerability to errors during the text recognition stage, particularly when dealing with low-quality scans. These errors can substantially compromise the overall performance of the subsequent Natural Language Processing (NLP) stage. 
In an effort to mitigate this challenge, the authors of \cite{prl22}  introduced  a transformer-based model to jointly perform text transcription and named entity recognition from the document images without an intermediate HTR stage. 
This innovative approach initiates by utilizing a CNN to extract visual features from the input images. Subsequently, these features are fed into a transformer encoder to generate hidden representations. Finally, a decoder is employed to transform these representations into a sequence of transcribed characters and named entity tags. 

Recent advancements in document understanding have witnessed the emergence of end-to-end models that seamlessly integrate image and text data, offering a holistic approach to document understanding within multimodal architectures. 
For example, the LayoutLM variants~\cite{ref_model6,ref_model7,ref_model8}, UDoc~\cite{ref_model15}, and UDOP~\cite{ref_model16}, go beyond mere visual processing. Take, for instance, LayoutLM in its versions 2 \cite{ref_model7} and 3 \cite{ref_model8}, which incorporates a spatial-aware self-attention mechanism into the Transformer architecture. This enhancement allows the model to gain a deeper understanding of the relative positional relationships among various text blocks. These models, often pre-trained in a self-supervised manner, exhibit remarkable performance in a spectrum of downstream tasks. These tasks include form understanding, receipt comprehension, document visual question answering, document image classification, and document layout analysis.

Similarly, the groundbreaking DONUT (DOcumeNt Understanding Transformer) model \cite{donut}, introduced a single-stage approach to document understanding. DONUT encompasses the analysis of document layouts to detect writing areas, text recognition using a lexicon of subwords, and named entity detection using specific TAGs, bolstered by a powerful external language model (BART). DONUT is pre-trained on synthetic documents, with ground truth provided as a sequence of subwords and TAGs, eliminating the need for segmentation ground truth. This paradigm shift simplifies Document Understanding to a task of learning a tagged language, assuming that the system possesses vision capabilities to construct high-level visual representations. In a similar vein, Adobe introduced DESSURT \cite{Dessurt}, following a comparable approach to integrate OCR and entity recognition. Additionally, the Pix2struct architecture \cite{peter} falls into the same category of integrated systems for document understanding. These innovations mark significant strides in document analysis streamlining and underline the evolving landscape of document understanding methodologies. However, the effectiveness of an invoice processing approach still depends on the quality and diversity of available data. In the following section, we provide a list of existing invoice datasets.

\subsection{Existing Datasets }
In this section, we shift our focus towards a comprehensive exploration of existing datasets tailored specifically for invoice document images analysis and understanding. Within this domain, researchers have access to some document image datasets, as outlined in Table \ref{tab1}. Nevertheless, it is vital to emphasize that only a select few of these datasets are notably well-suited for information extraction from invoices.

For example, consider the IIT-CDIP dataset \cite{ref_data1} and its subset, RVL-CDIP \cite{ref_data2}. These datasets provide extensive image collections, covering a diverse array of document types, including invoices. However, it is important to acknowledge that these images commonly exhibit noise and may have relatively lower resolutions. Furthermore, the data set only includes document class labels for document image classification, lacking crucial bounding-box coordinates and textual details necessary for a more comprehensive analysis.

The SROIE dataset \cite{ref_data3} contains a set of 1,000 scanned receipt images, each accompanied by its respective annotations. Some receipts exhibit complex layouts, reduced quality, low scanner resolutions, and scanning distortions, making the data set more challenging. Another receipt dataset is CORD dataset \cite{ref_data6}, featuring an extensive collection of more than 11,000 fully annotated images depicting Indonesian receipts.
However, it's important to acknowledge that the SROIE and CORD datasets may exhibit limited diversity in terms of layouts, as many of the receipts follow similar structural patterns. Compared to receipts, invoices present distinct challenges for information extraction due to their diverse layouts and structured data presentation. Invoices commonly feature complex elements, such as tables, as well as multiple sections encompassing billing information, shipping details, and itemized lists. Given these differences, the SROIE and CORD datasets may not be the most suitable choice for analyzing and understanding invoice document images.

Furthermore, the FUNSD data set \cite{ref_data5} offers a collection of 199 images in scanned form that feature diverse layouts and varying levels of noise. Remarkably, the dataset boasts extensive and accurate annotations, furnishing valuable ground-truth information for form understanding in noisy scanned documents. However, it is essential to recognize that the dataset's relatively modest size poses a notable limitation, which may be challenging for certain machine learning models that require a larger and more diverse dataset to be trained. Furthermore, it's essential to note that the forms in the FUNSD dataset span diverse fields, such as marketing, advertising, and scientific reports, introducing a distinct layout variety that distinguishes them from typical invoice formats.
\newline
Existing invoice datasets, while valuable for model training and evaluation, often suffer from significant limitations. These issues primarily revolve around limited diversity, with datasets frequently skewed towards specific domains or regions. Such biases can lead to reduced accuracy when applying models to invoices from various contexts. Furthermore, an imbalanced data distribution is a common challenge, where some data sets contain a disproportionate number of a certain type of invoices, affecting model performance. Additionally, the lack of variability in invoice formats within datasets can hinder a model's adaptability to different layouts and structures. These challenges underscore the pressing need for more comprehensive and inclusive datasets to enhance the robustness and effectiveness of invoice processing models.

To tackle these challenges, we introduce FATURA, a multi-template invoice document images dataset. This dataset exhibits a remarkable diversity in invoice layouts, aiming to significantly enhance the capabilities of models for analyzing and understanding unstructured documents. Our aim is to enhance the performance of these models, ensuring their effectiveness in real-world applications.

\begin{table}[]
\label{tab1}
\begin{tabular}{c|c|c|c|c}
                  & \textbf{Detailed Annotations} & \textbf{Large Size} & \textbf{Diverse} & \textbf{Contains Invoices} \\ \hline
\textbf{IIT-CDIP} & No                            & Yes (7m)                & Yes              & Yes                        \\
\textbf{RVL-CDIP} & No (document-level labels)    & Yes (400k)               & Yes              & Yes                        \\
\textbf{SROIE}    & Yes                           & No  (1k)                & No               & No (receipts)              \\
\textbf{FUNSD}    & Yes                           & No  (199)                & Yes              & Yes                        \\
\textbf{CORD}     & Yes                           & Yes (+11k)                & No               & No (receipts)             
\caption{Summary of existing invoice-related datasets}
\end{tabular}
\end{table}

\section{FATURA dataset}
In this section, we introduce the FATURA dataset, illustrated in Fig. \ref{fig:samples}. We describe the generation process of the documents, their structure and their content. The database containing both the images and the annotations is freely available at \url{https://zenodo.org/record/8261508}.

\subsection{Invoice Generation Process}
The generation of synthetic invoice templates is a complex yet crucial step in creating a diverse and representative dataset for document analysis and understanding. This process involves multiple stages, each carefully designed to strike a balance between realism and diversity while addressing crucial privacy concerns.
The generation process revolves around the transformation of a blank canvas into plausible invoice templates. This intricate procedure consists of several key steps aimed at producing diverse and representative templates.

The first step is to gather a comprehensive dataset of real invoice images. These genuine invoice templates serve as the fundamental building blocks for our synthetic templates. To ensure that the generated templates encapsulate essential components, such as buyer information, total amounts, and invoice dates, the templates are carefully selected based on their content.

Secondly, each real image is annotated using the VGG Image Annotator tool. The primary objective of this annotation process is to establish a well-annotated layout for each invoice template. At this stage, the textual content is omitted because it will be generated randomly in subsequent phases. Our focus is on capturing the structural blueprint of the invoices, including the placement and dimensions of different elements.   
We also incorporate logos into our synthetic templates to enhance their realism. A unique logo is generated for each template using a pre-trained open-source Text-to-Image Latent Diffusion model~\cite{ref_model9}.  It's important to note that all images from the same template share a common background color, adding to the visual coherence of the dataset.

Generating variants of each invoice template is a key step in enriching the dataset. The algorithm begins with a blank canvas and proceeds to insert the relevant  textual information  in each component of the original template. For each component, a bounding box is created according to the original coordinates, preserving the spatial arrangement of the elements. The text generation process is carefully tailored to create invoices that mimic real-world diversity. To further enrich the dataset's realism and variability, certain components such as sender/receiver names, addresses, and product descriptions are populated with randomly selected text from a predefined repository of plausible texts. The algorithm also verify that the total amount on the invoice matches the sum of all amounts of individual products. 


\begin{figure}[h]
    \centering
    \includegraphics[width=12cm]{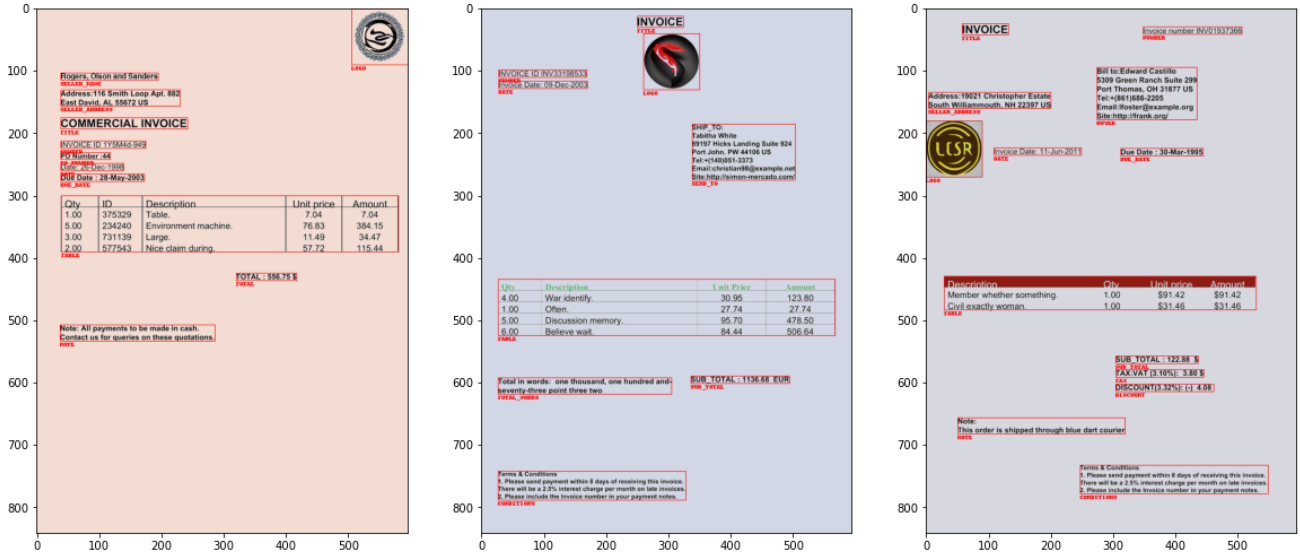}
    \caption{Examples of annotated images from different templates}
    \label{fig:samples}
\end{figure}

\subsection{Dataset Description}
The dataset consists of 10000 JPEG images each accompanied by their corresponding JSON annotation files. These images are generated based on a set of 50 distinct templates. Templates cover a wide spectrum of designs, including variations in font styles, text placements, and graphical elements. This diversity reflects real-world scenarios and ensures the relevance of the data set to a wide range of applications. It's noteworthy that even within the same template, each generated invoice image exhibits distinct textual content, further enhancing the dataset's realism and utility.

Annotations for the FATURA dataset are available in three different formats to accommodate various research needs. The first format adheres to the well-established COCO annotation format, widely recognized and utilized in computer vision tasks. The second format is meticulously tailored for seamless integration with the HuggingFace\footnote{https://huggingface.co} Transformers library~\cite{translibref}, a renowned resource in the field. This format is specifically designed to align with the capabilities of the LayoutLMv3 architecture. The third format represents our dataset's standard annotation format, which can be used in research contexts where a custom or unique annotation schema is preferred.

\begin{figure}[h]
    \centering
    \includegraphics[width=12cm]{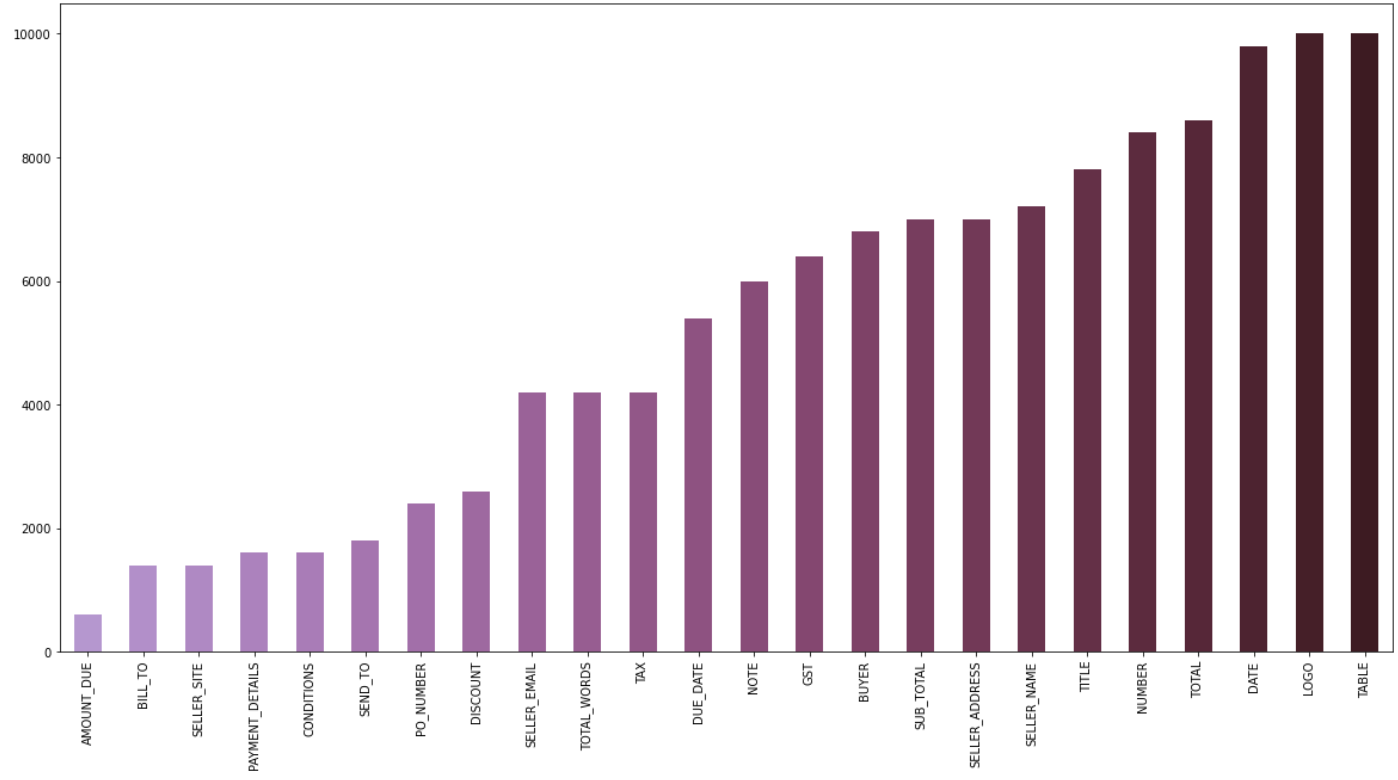}
    \caption{Class occurrence distribution in the FATURA Dataset}
    \label{fig:counts}
\end{figure}

We have identified $24$ distinct classes corresponding to the different fields that can be extracted from an invoice, as indicated in Table \ref{tab:counts}. It should be noted that the frequency of these classes exhibits considerable variation, leading to an imbalanced dataset, as visually depicted in Figure \ref{fig:counts}.

\begin{table}[!ht]
    \centering
    \resizebox{\textwidth}{!}{\begin{tabular}{|c|c|c|c|c|}
    \hline
    Class/Text-Region &  n°Instances & \% among objects & n°Templates & Description\\
    \hline
    TABLE & 10000 & 8\% & 50 & Information about bought products \\
    LOGO & 10000 & 8\% & 50 & Template logo \\
    DATE & 9800 & 7.9\% & 49 & Date of purchase \\
    NUMBER & 8800 & 7\% & 44 & Invoice ID \\
    SELLER ADDRESS & 8200 & 6.6\% & 41 &  \\
    TOTAL & 8000 & 6.4\% & 40 & Total amount after tax and discount \\
    TITLE & 7400 & 5.9\% & 37 & \\
    SELLER NAME & 6800 & 5.5\% & 34 & \\
    SUB-TOTAL & 6800 & 5.5\% & 34 & Total amount before tax and discount \\
    BUYER & 6200 & 5\% & 31 & Buyer information \\
    DUE DATE & 5800 & 4.7\% & 29 & \\
    NOTE & 5200 & 4.2\% & 26 & Remarks and footers \\
    GST & 5000 & 4\% & 16 & Goods and services tax \\
    SELLER EMAIL & 4400 & 3.5\% & 22 & \\
    TOTAL WORDS & 4000 & 3.2\% & 20 & Total in words \\
    TAX & 3800 & 3\% & 19 & \\
    PAYMENT DETAILS & 2600 & 2\% & 13 & Bank information \\
    DISCOUNT & 2400 & 1.9\% & 12 & \\
    CONDITIONS & 2400 & 1.9\% & 12 & Payment terms and conditions \\
    SEND TO & 1800 & 1.4\% & 9 & To whom the invoice is sent \\
    BILL TO & 1600 & 1.3\% & 8 & To whom the bill is sent \\
    PO NUMBER & 1400 & 1.1\% & 7 & Purchase order number\\
    SELLER SITE & 1400 & 1.1\% & 7 & Website of the seller \\
    AMOUNT DUE & 800 & 0.6\% & 4 & Total amount to be paid \\ \hline
    \end{tabular}}
    \caption{Description of the information contained in the different invoices}
    \label{tab:counts}
\end{table}

\subsection{Evaluation Strategies}
In this section, we present two distinct evaluation strategies that facilitate the training and assessment of models on our dataset, each offering unique insights into model performance. 

In the first evaluation strategy, we employ an intra-template-centric scenario. For each of the 50 distinct templates, the generated images are randomly partitioned into three subsets: training, validation, and testing. In this scenario, models are trained on all templates, thereby ensuring exposure to various layouts and styles. Consequently, during testing, models encounter new images based on familiar templates. This approach provides a valuable assessment of a model's ability to generalize its understanding of document layouts and adapt to diverse content.

The second evaluation strategy adopts an inter-template centric perspective, emphasizing the diversity of templates and layouts. Here, the templates are randomly segregated into a training set, and a distinct set of templates is reserved for validation and testing. This evaluation scenario evaluate the models' performance on different unseen templates/layouts, rather than the same templates with different content.

By embracing these two complementary evaluation strategies, we ensure a comprehensive assessment of the performance of the model. The first strategy focuses on the adaptability of models to diverse content within familiar templates, while the second strategy challenges models to generalize effectively across a variety of templates and layouts, evaluating their adaptability to new, unseen document structures. 
This approach facilitates a holistic evaluation of models in the context of real-world document analysis and understanding tasks.


\section{Experimental results}
In this section, our objective is to establish comprehensive benchmarks by training and testing models on a spectrum of document layout analysis and document understanding tasks. We meticulously evaluate distinct approaches:

\begin{itemize}
\item Visual-based approach: In this approach, we leverage object detection techniques to classify text regions within the document images. This method relies solely on visual cues and layout information to analyze and categorize document content.

\item Multi-Modal approach: In contrast, our second approach employs a multi-modal strategy for token-level classification. Here, we integrate both visual and textual information to classify and understand document content at a more granular level.

\item Hybrid approach: Additionally, we explore the synergies between these approaches by combining object detection and token classification methods. This hybrid end-to-end approach aims to harness the strengths of both techniques, offering a potentially powerful solution for document analysis and understanding tasks.
    
\end{itemize}

By evaluating these diverse strategies, we aim to provide a comprehensive overview of model performance and shed light on the most effective methods for various document-related challenges.

\subsection{ Visual-Based Approach }

This approach consists in training an object detection model designed to locate and classify entire text regions within the document images.
We employ the YOLOS (You Only Look At One Sequence)~\cite{ref_model11}. YOLOS  is an object detection architecture built upon the foundation of the vanilla Vision Transformer~\cite{ref_model12}. This unique architecture has demonstrated performance levels comparable to other state-of-the-art object detection approaches that harness the combined power of the Transformer architecture and Convolutional Neural Networks~\cite{ref_model13}, such as DETR~\cite{ref_model14}. To adapt YOLOS for our specific task, we fine-tune the YOLOS-Ti version (the smallest version of YOLOS), pre-trained on COCO 2017~\cite{ref_coco}. 
The model takes as input a sequence of flattened image patches, followed by one hundred randomly initialized detection tokens. The output tokens corresponding to the detection tokens are subsequently processed by a multilayer perceptron head to generate boxes and associated classes.

\begin{table*}[]
    \centering
    \begin{tabular}{|c|c|c|c|}
        \hline
         Metric & Training & Validation & Test \\ \hline
         mAP@IOU=50 & 97.57\% & 97.77\% & 97.57\% \\ \hline
         mAP@IOU=75 & 81.92\% & 81.31\% & 81.84\% \\ \hline
         mAR@maxDets=10 & 78.05\% & 77.55\% & 77.73\% \\ \hline
         mAR@maxDets=15 & 78.08\% & 77.56\% & 77.75\% \\ \hline
        
    \end{tabular}
    \caption{YOLOS Results: First Evaluation Strategy}
    \label{tab:yolos_res1}
\end{table*}

\begin{table*}[]
    \centering
    \begin{tabular}{|c|c|c|c|}
        \hline
         Metric & Training & Validation & Testing \\ \hline
         mAP@IOU=50 & 99.1\% & 43.6\% & 43.58\% \\ \hline
         mAP@IOU=75 & 97\% & 32.66\% & 32.7\% \\ \hline
         mAR@maxDets=10 & 90.17\% & 32.14\% & 32.14\% \\ \hline
         mAR@maxDets=15 & 90.2\% & 32.14\% & 32.15\% \\ \hline
         
    \end{tabular}
    \caption{YOLOS Results: Second Evaluation Strategy}
    \label{tab:yolos_res2}
\end{table*}
In Table~\ref{tab:yolos_res1}, we present the results for the first evaluation strategy (intra-template evaluation), while Table~\ref{tab:yolos_res2} displays the results for the second evaluation strategy (inter-template evaluation). The results include mean Average Precision (mAP) scores at a specific IOU value. Additionally, we report the averaged maximum recall (mAR) which represents the maximum recall given a fixed number of detections per image, averaged over categories and IoUs.   

The results obtained from our experimentation reveal a notable trend: the first evaluation strategy, which emphasizes intra-template evaluation, consistently outperforms the second evaluation strategy focused on inter-template assessment. The central factor that contributes to the observed performance disparity lies in the inherent capabilities of the YOLOS model. In fact, YOLOS, by design, excels at analyzing the layout and structure of document images and succeeds in identifying and categorizing complete text regions within the documents based on visual cues, rather than comprehending textual content.
In the first evaluation strategy, where the model encounters new images based on the same templates it has been trained on, the model showcases its strength in intra-template understanding. It has learned the intricate layouts and text region placements specific to each template during training. Consequently, when tasked with analyzing new instances of familiar templates during testing, the model excels in accurately localizing and classifying text regions. 
On the contrary, the second evaluation strategy challenges the model with templates that it has not seen during training. While the model has a strong foundation in layout analysis, it faces difficulties when confronted with entirely new templates that deviate from the ones it has been exposed to. This inter-template generalization proves to be a more demanding task, as it requires the model to adapt to diverse layouts, structures, and visual cues that may not align with its training data.

These results have significant implications for the application of detection-based models in document analysis. Models like YOLOS, tailored for layout understanding and object detection, excel at tasks related to visual recognition and structuring, but do not possess text-completion capabilities. Therefore, their performance is highly dependent on the familiarity of the templates. In conclusion, these results highlight the need for a nuanced approach  equipped with text comprehension capabilities like LayoutLMv3. 

\subsection{Multi-Modal LayoutLMv3-Based Approach}

In our second approach, we harness the power of the LayoutLMv3 architecture to create a multi-modal framework capable of modeling the intricate interactions among text, layout, and image components within document images. This approach represents a significant departure from the purely visual-based strategy discussed earlier, as it embraces a holistic understanding of documents by integrating textual, layout, and visual information.
LayoutLMv3, an evolution of its predecessors, is a state-of-the-art architecture designed explicitly for document understanding tasks. It combines the strengths of the Transformer architecture with multi-modal capabilities, allowing it to process text, layout, and image information seamlessly. This versatility positions LayoutLMv3 as a powerful tool for modeling complex document structures, making it well-suited for a wide range of document analysis and understanding tasks.

Although good performance has been achieved with LayoutLMv3, domain knowledge of one document type cannot be easily transferred to another. In addition, one significant drawback of LayoutLM is its reliance on a complex preprocessing step for word-bounding-box segmentation. Document images like invoices featuring a complex and varied layout (key-value pairs in a left-right layout, tables in a grid layout, etc.) often need to undergo extensive preprocessing to identify and delineate individual word bounding boxes accurately. This step can be computationally expensive and may require additional expertise in data preparation, making the model less accessible for users without specialized knowledge. 

On the other hand, LayoutLM's performance is heavily reliant on the quality of Optical Character Recognition (OCR) results. Inaccuracies or errors in text recognition by the OCR system can negatively impact the model's ability to understand document content. LayoutLMv3 assumes that OCR provides precise and complete text transcriptions, which may not always be the case, especially when dealing with complex documents like invoices. 

In our specific scenario, we attempted to train and deploy LayoutLMv3 at the word level, utilizing the FATURA dataset. Regrettably, the outcomes fell short of expectations, with the model struggling to precisely identify and delineate individual word bounding boxes. This posed a significant challenge to the model's performance in accurately processing the dataset.  

For these reasons, we decided to apply LayoutLMv3 at the region level instead of the word level for key-value extraction in the context of invoices. This can help the model capture the contextual relationships between these regions and the associated key-value pairs. In addition, invoices can have diverse layouts, with variations in the placement of key information. By analyzing regions as a whole, LayoutLMv3 can adapt to different layout styles, making it more robust in handling invoices from various sources and formats. Working at the region level can also reduce the noise introduced by irregular spacing, formatting, or variations in text sizes within key-value pairs. The model can focus on the overall structure and content of each region, improving the accuracy of key-value extraction.

Before delving into the results of our hybrid approach, it is essential to establish a baseline reference approach. In this reference approach, we utilize region-bounding box annotations and their corresponding text ground-truth transcriptions during the inference stage. However, it is important to note that this approach is not practically viable and is primarily intended for comparative purposes. Its performance represents the upper bound as it leverages ground-truth annotations, which are typically unavailable in real-world scenarios. The purpose of introducing this reference approach is to assess and contextualize the performance of our hybrid approach, which operates under more realistic conditions. To evaluate the model performance, we report the precision, recall, and F1-score metrics.



\begin{table*}[]
\centering
\begin{tabular}{|c|c|c|c|}
\hline
Split      & F1-Score  & Recall & Precision \\ \hline
Training   & 100\%        & 100\%             & 100\%         \\ \hline
Validation & 95.7\%     & 97.5\%       & 95.7\%      \\ \hline
Testing    & 95.7\%     & 97.5\%        & 95.7\%      \\ \hline
\end{tabular}
\caption{\label{tab:layoutlm_res} LayoutLMv3 Results: Second evaluation strategy}
\end{table*}

In the context of the first evaluation strategy (intra-template evaluation), the model shows high performance, achieving remarkable $99\%$ across all metrics for both the validation and testing sets. However, when transitioning to the second evaluation strategy, the model performance exhibits a slight decrease, as indicated in Table~\ref{tab:layoutlm_res}. This expected decline in performance can be attributed to the increased complexity of the second scenario, which involves diverse templates and layouts present across distinct sets.

\subsection{Hybrid  Approach}
This approach combines the strengths of YOLOS object detection and the multimodal LayoutLMv3 model to achieve precise extraction of key value information within text regions.

In the initial phase, we employ the pre-trained YOLOS model from our earlier experiment to identify text regions. Although YOLOS, like other object detection architectures, allows for overlapping bounding boxes, our generated dataset features non-overlapping text regions. To address this, we implement a modified Non-Maximum Suppression technique with an IOU threshold of $30\%$ to eliminate redundant bounding boxes.

Subsequently, an OCR system, specifically EasyOCR~\cite{ref_model15}, is used to extract text content from the cropped regions. The resulting bounding boxes and their corresponding text content are then input into the LayoutLMv3 model for token classification. The model assigns labels to individual tokens, and the most frequently occurring label (mode) is assigned as the label for the entire text region. This comprehensive approach ensures accurate and context-aware classification within the identified text regions.

This approach consists of using YOLOS object detection to detect text regions, followed by the multi-modal LayoutLMv3 model that classifies individual tokens, rather than text regions.
We use the pre-trained YOLOS model from the previous experiment.

First, the YOLOS model is used to localize text-regions. The YOLOS architecture, just like other object detection architectures, allows for overlapping boxes. As the text regions in our generated dataset do not overlap, we deal with this issue by removing certain bounding boxes using a variant of Non-Maximum Suppression with an IOU threshold of 30\%.

Then, an OCR system (we used EasyOCR~\cite{ref_model15}) is used to extract text from the cropped regions. The bounding boxes and the corresponding texts are then fed to the LayoutLMv3 model to get token classification labels. Finally, the most frequent label is used as the label of the entire text-region.
\begin{figure}[h]
    \centering
    \includegraphics[width=12cm]{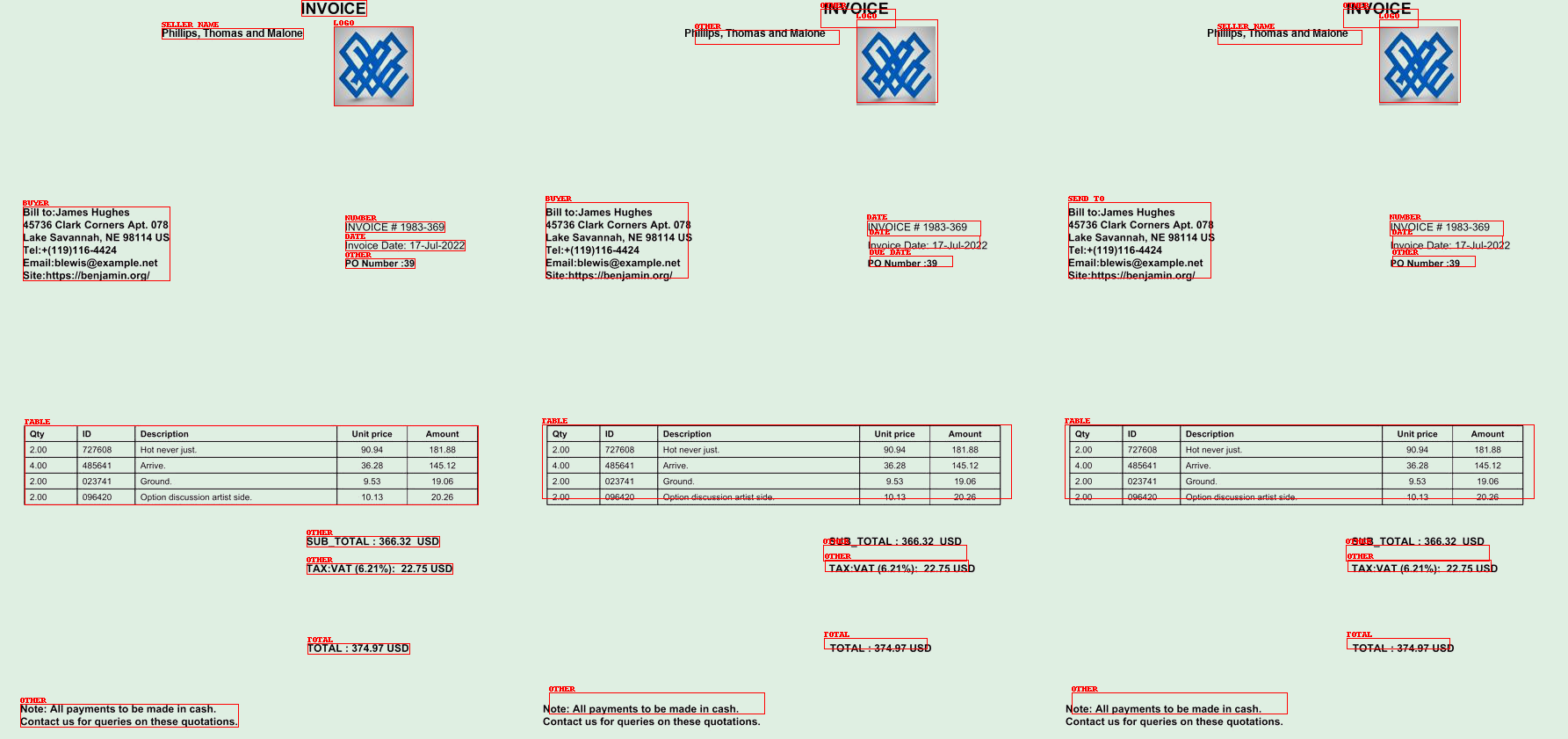}
    \caption{Comparison between ground-truth (left), YOLOS predictions (center), and hybrid approach predictions (right)}
    \label{fig:samples}
\end{figure}

\begin{figure}[h]
    \centering
    \includegraphics[width=12cm]{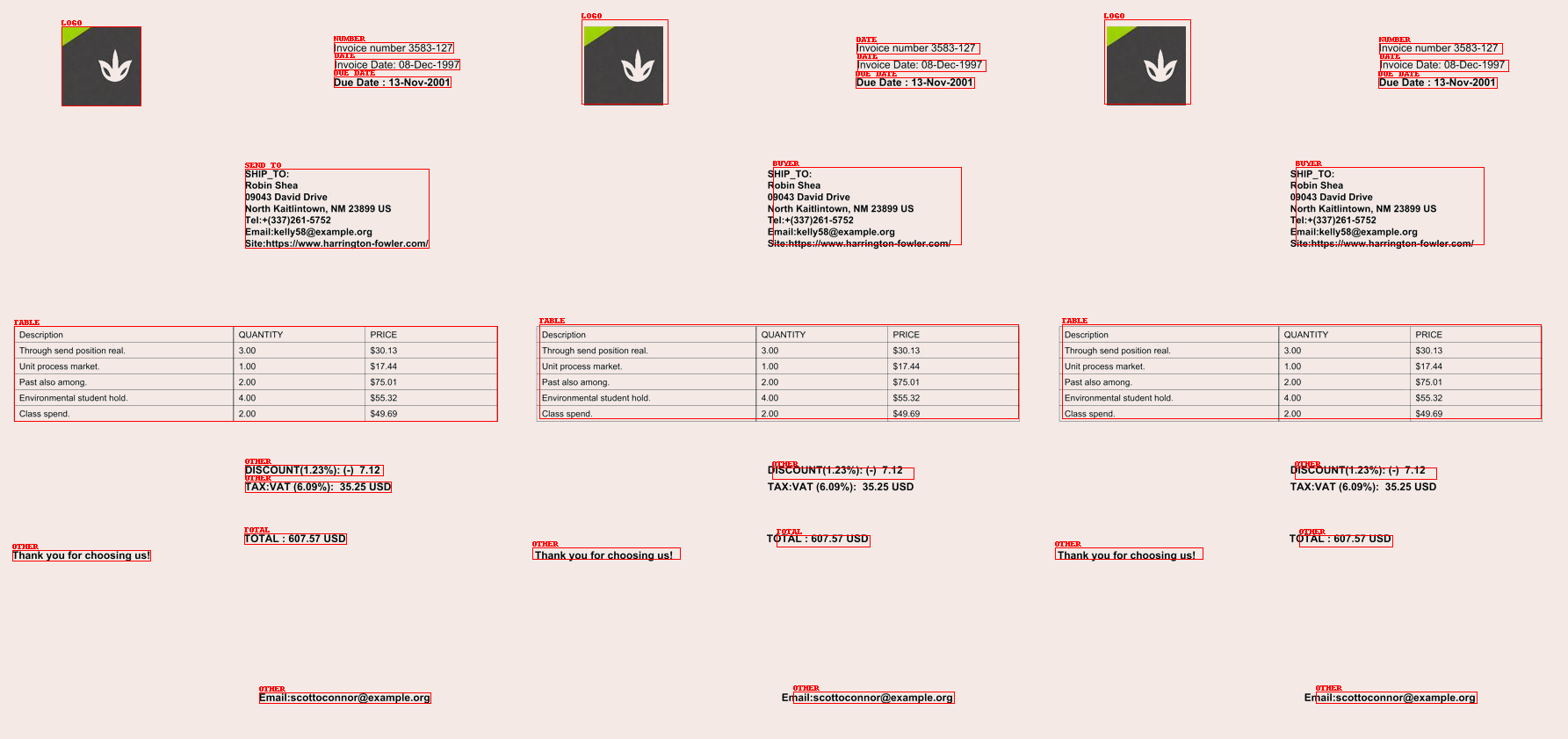}
    \caption{Comparison between ground-truth (left), YOLOS predictions (center), and hybrid approach predictions (right)}
    \label{fig:samples}
\end{figure}

\begin{figure}[h]
    \centering
    \includegraphics[width=12cm]{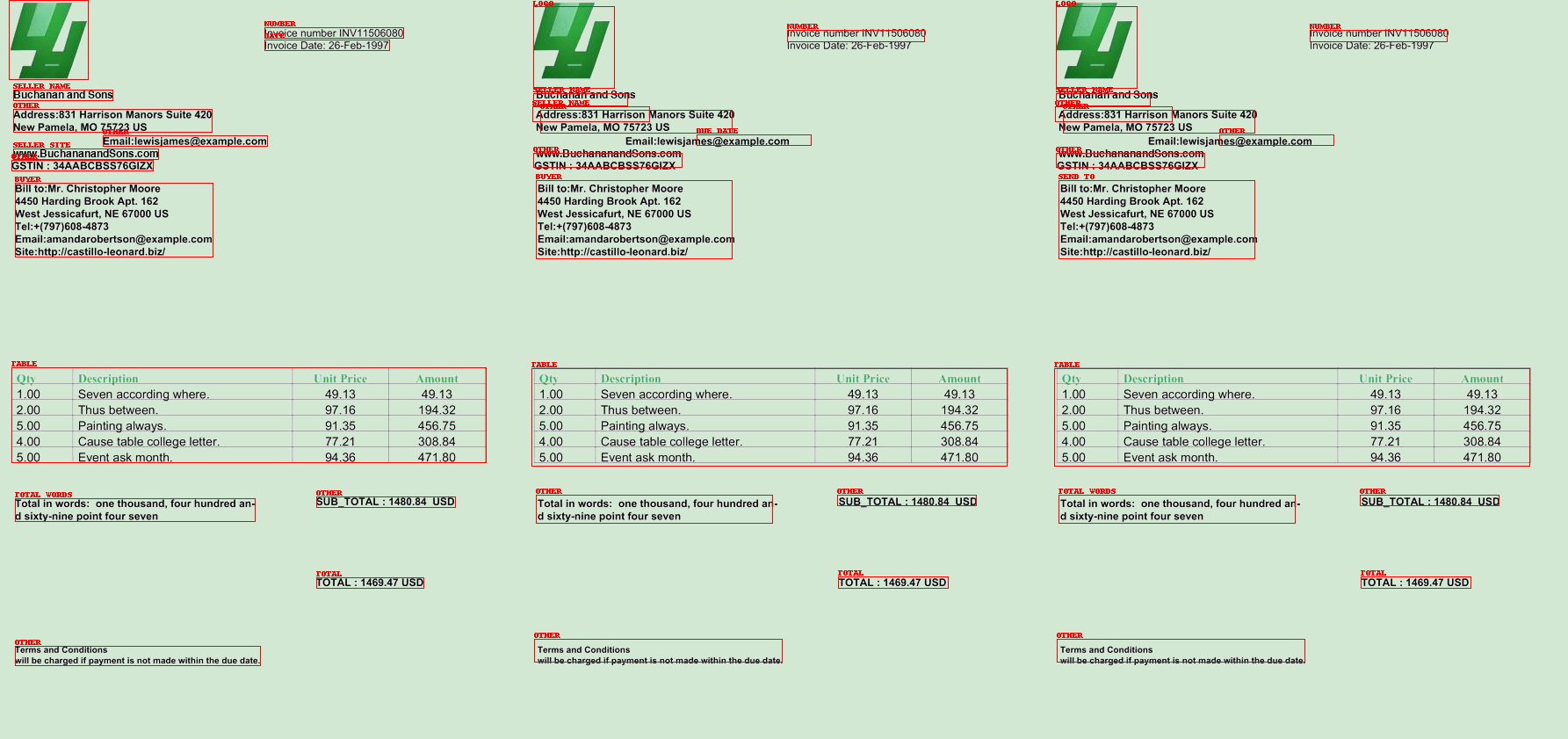}
    \caption{Comparison between ground-truth (left), YOLOS predictions (center), and hybrid approach predictions (right)}
    \label{fig:samples}
\end{figure}

\begin{figure}
    \centering
    \includegraphics[width=12cm]{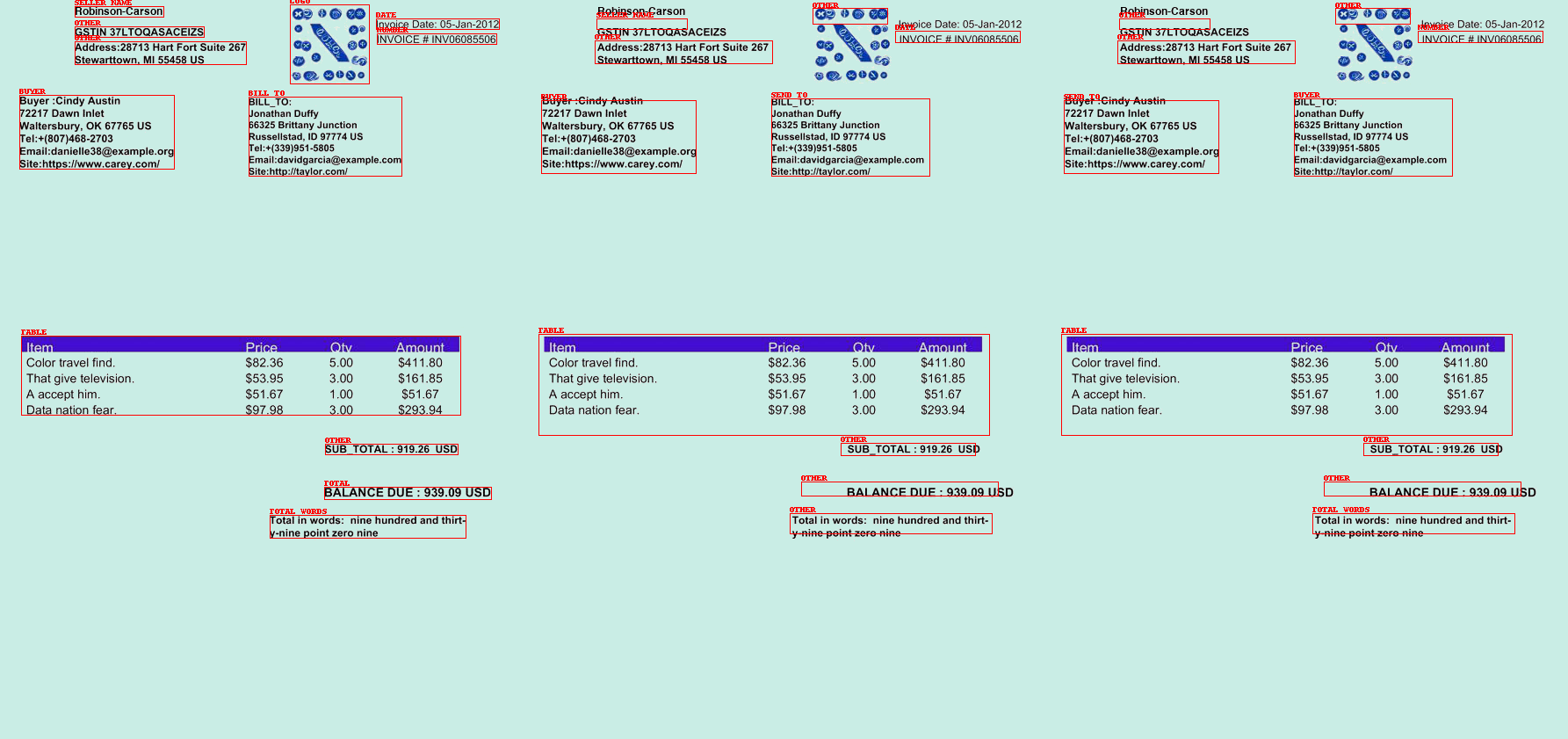}
    \caption{Comparison between ground-truth (left), YOLOS predictions (center), and hybrid approach predictions (right)}
    \label{fig:samples}
\end{figure}


\begin{table*}[]
\centering
\begin{tabular}{|c|c|c|c|c|}
\hline
Split      & F1-Score  & Recall & Precision \\ \hline
Training   & 88\%        & 85\%      & 92\%         \\ \hline
Validation & 88\%         & 85\%   & 92\%      \\ \hline
Testing    & 88\%         & 84\%   & 92\%      \\ \hline
\end{tabular}
\caption{\label{tab:hybrid1_res} Hybrid Approach Results: First Evaluation Strategy}
\end{table*}


\begin{table*}[]
\centering
\begin{tabular}{|c|c|c|c|c|}
\hline
Split      & F1-Score  & Recall & Precision \\ \hline
Training   & 79.3\%            & 70.8\%      & 91.1\%         \\ \hline
Validation & 56.3\%          & 47.7\%   & 70.7\%      \\ \hline
Testing    & 56.3\%         & 47.7\%   & 70.7\%      \\ \hline
\end{tabular}
\caption{\label{tab:hybrid2_res} Hybrid Approach Results: Second Evaluation Strategy}
\end{table*}
 
Similar to the preceding approaches, our evaluation of the hybrid approach, as depicted in Table \ref{tab:hybrid1_res}, reveals a notable performance advantage in the context of the first evaluation scenario compared to the outcomes in the second evaluation strategy, outlined in Table \ref{tab:hybrid2_res}. 

When contrasting these results with those of the previous section, which employed region bounding box annotations and corresponding text ground truth transcriptions during inference, we observe a discernible performance gap. This variance can be attributed to inherent segmentation errors in the YOLOS model, coupled with OCR inaccuracies, both of which exert a detrimental impact on the model's capacity to accurately comprehend the document content. These errors introduce challenges that affect the model's ability to precisely classify tokens within text regions, particularly in the more complex and diverse context of the second evaluation strategy.

The effectiveness of our hybrid approach in accurately identifying specific document fields, when compared to the purely-visual approach, can be attributed to its ability to harness both visual and textual information. By incorporating both modalities, the second approach gains a holistic understanding of the document content, allowing it to make more informed decisions during the segmentation process. This synergy between visual and textual data empowers the second approach to excel in situations where the first approach, limited to visual cues alone, may falter. For example, in figures 3 and 4, one can see the YOLOS model attributing wrong labels to the date and invoice id fields, whereas the hybrid model, augmented with textual input, doesn't fall into such confusion. The visual model might occasionally not associate a label to an important region by attributing it to the "other" class. This can be seen in figures 5 and 6 in the total words region, which the hybrid model correctly identifies. Although there are exceptions where the hybrid models misidentifies region that are correctly classified by the visual model, such as in image 5 where the receiver ("send to") was misidentified as the buyer, the hybrid model surpasses the visual one in most cases.

\section{Conclusion}
In summary, this paper has made significant strides in advancing the domain of document analysis and understanding through innovative approaches and the introduction of a valuable resource. We presented the FATURA dataset, a diverse collection of multi-layout and annotated invoice documents, addressing the critical need for high-quality, unstructured document datasets, with a particular focus on invoices that are vital in numerous real-world applications.

Throughout our investigation, we thoroughly examined various evaluation strategies and employed state-of-the-art models such as LayoutLMv3 and YOLOS to tackle intricate document analysis tasks. We introduced our hybrid approach, harnessing the synergy between object detection and token-level classification to enhance document understanding, even in the face of segmentation and OCR inaccuracies.

In conclusion, our contributions in dataset creation, model evaluation, and the proposed hybrid approach establish a robust foundation for future research in document analysis and understanding. We believe that the FATURA dataset and the insights presented in this paper will serve as valuable resources, inspiring further innovations and advancements in the field. One promising perspective is the extension of this dataset to encompass multi-lingual invoices, catering to a broader range of document types and languages. This expansion holds great potential for advancing the field and addressing the evolving demands of real-world document analysis applications.
\section{Conflict of Interest}
The authors declare that there is no conflict of interest regarding the publication of this paper.

\section{Data Availability}
The dataset is accessible at this \href{https://zenodo.org/record/8261508}{URL\footnote{https://zenodo.org/record/8261508}}.

\bibliography{sn-bibliography}

\end{document}